\documentclass[11pt]{article}
\usepackage{ranlp2011}
\usepackage{times}
\usepackage{url}
\usepackage{latexsym}
\usepackage[utf8]{inputenc}
\usepackage{textcomp}
\usepackage{amsfonts}
\usepackage{float}
\usepackage{qtree}
\floatstyle{boxed} 
\restylefloat{figure}

\title{Tree Transducers, Machine Translation, and Cross-Language Divergences}

\author{Alex Rudnick\\
  School of Informatics and Computing, Indiana University \\
  Bloomington, Indiana, USA\\
  {\tt alexr@cs.indiana.edu} }
\date{}

\begin{document}
\maketitle

\begin{abstract}
Tree transducers are formal automata that transform trees into other trees.
Many varieties of tree transducers have been explored in the automata theory
literature, and more recently, in the machine translation literature.  In this
paper I review T and xT transducers, situate them among related formalisms, and
show how they can be used to implement rules for machine translation systems
that cover all of the cross-language structural divergences described in Bonnie
Dorr's influential article on the topic. I also present an implementation of xT
transduction, suitable and convenient for experimenting with translation rules.
\end{abstract}

\section{Introduction}
Word-based approaches to statistical machine translation, starting with the
work from IBM in the early 1990s \cite{DBLP:journals/coling/BrownPPM94} have
been successful both in use in production translation systems and in
invigorating MT research. Since then, newer phrase-based MT techniques such as
the alignment template model \cite{DBLP:journals/coling/OchN04}, and
hierarchical phrase-based models \cite{chiang:2005:ACL} have made significant
improvements in SMT translation quality.

Despite their sophistication and apparent complexity, many word-based and
phrase-based SMT models can be implemented entirely in terms of finite-state
transducers. This allows researchers to make use of the rich automata
literature for finding clean and efficient algorithms; it is also useful from a
software engineering perspective, making it possible to do experiments quickly,
using generic toolkits for programmatically manipulating finite-state
transducers. Several such packages are freely available, such as OpenFST
\cite{openfst} and the WRTH FSA Toolkit \cite{kanthak-ney:2004:ACL}.

However, since they make no attempt to explicitly model the syntax of either
involved language, and typically use simple n-gram models to guide generation,
the output of word-based SMT systems can be syntactically incoherent,
especially in light of long-distance dependencies. Additionally, word-based SMT
models have difficulties encoding word order differences across languages. 

So we have seen new methods in MT that explicitly model syntax, where typically
the grammar of a language, and the relationships between the grammars of two
languages, can be learned from treebanks. There are many different available
theoretical frameworks for describing syntax and transformations over syntactic
representations. Both from a theoretical standpoint, and as MT implementors, we
would like a framework that is clean and general, and is suitably expressive
for explicitly capturing syntactic structures and the divergences across
languages. We would also like one for which there are efficient algorithms for
training rules and performing transduction (i.e., decoding at translation
time), and ideally one for which a good software toolkit is freely available.

Not all syntactic relationships can be cleanly represented with every syntactic
formalism; each formalism has its own expressive power. Bonnie Dorr provides us
with an excellent test bed of seven cross-language divergences that may occur
when we want to perform translation, even between languages as closely related
as English, Spanish and German \cite{DBLP:journals/coling/Dorr94}. While these
divergences do not totally describe the ways in which languages can differ in
their typical descriptions of an event, they provide a concrete starting point,
and are easily accessible.

In this paper, I specifically investigate T and xT transducers, situate them in
the space of formalisms for describing syntax-based translation, and
demonstrate that xT transducers are sufficient for modeling all of the
syntactic divergences identified by Dorr. I also present \texttt{kurt}, a small
software toolkit for experimenting with these transducers, which comes with
sample translation rules that handle each of Dorr's divergences.

In the rest of this paper, we will discuss some relevant grammar and transducer
formalisms, including a more in-depth look at T and xT transducers; go through
the linguistic divergences discussed by Dorr and explain why they might cause
difficulties for MT systems; show how xT transducers can be used to address
each of these divergences; present the software that I have built; review some
of the related work that has informed this paper; and finally, suggest future
possible directions for work with tree transducer-based MT.

\section{Grammars and Transducers}
Here we contrast several kinds of formalisms over strings, trees, and pairs of
strings and trees; please see Figure \ref{glossary} for a glossary of different
kinds of automata and grammars that will be referenced in this paper.  A
grammar describes a single set of strings or trees, and consists of a finite
set of rules that describes those strings or trees. Familiar formalisms for
grammars that describe sets of strings include context-free grammars and the
other members of the Chomsky Hierarchy. Some grammars describe sets of trees,
and these will be the main focus of the rest of this paper; when discussing
grammars over strings, I will specifically mention it.  For example, regular
tree grammars (RTG) is the class of grammars corresponding to context-free
grammars but describing trees; they describe the trees whose \textit{yield}
(string concatenation of the symbols at the leaves) is a context-free grammar
\cite{KnightGraehlOverview}.

Contrastingly, \textit{synchronous} grammars describe sets of pairs of objects;
here again, we are mostly concerned with synchronous grammars that describe
trees.  Formally, a synchronous grammar over trees establishes a mathematical
relation over two sets of trees, and allows us to answer the question of
whether, for a given pair of trees, that pair is in the relation. The
production rules of a synchronous grammar do not just describe one language,
but have pairs of production rules $<r_1,r_2>$, such that when $r_1$ is used to
derive a string in language $L_1$, $r_2$ must be used in the derivation of a
string in $L_2$.

\begin{figure*}
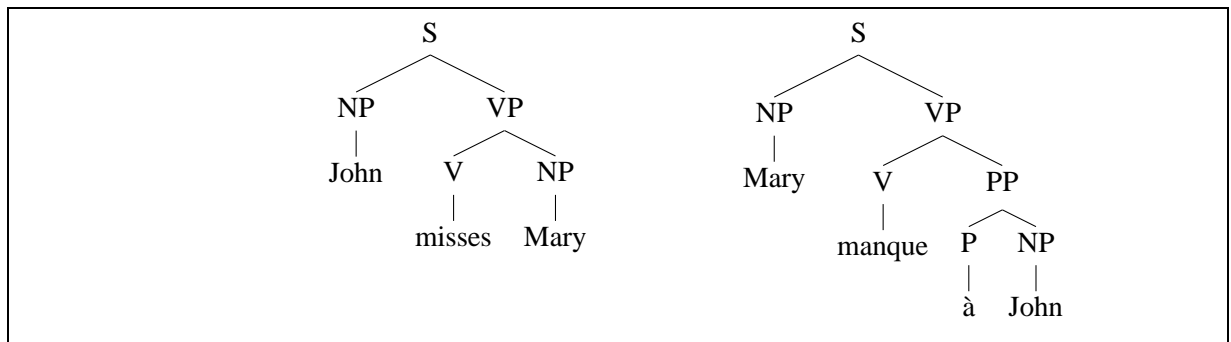

\begin{center}
\Tree [.S [.NP John ] [.VP [.V misses ] [.NP Mary ] ] ]
\Tree [.S [.NP Mary ] [.VP [.V manque ] [.PP [.P à ] [.NP John ] ] ] ]
\end{center}
\caption{Switching subject and object in translation to French. Example from
\cite{Chiang06anintroduction}. Also note the structural difference: there is a
PP subtree in French, not present in English.} 
\label{missesmary}
\end{figure*}

Thus synchronous grammars can be used for several kinds of tasks, such as
parsing parallel texts, generating parallel text, or most intuitively useful
for a machine translation setting, parsing text in one language while jointly
generating parse trees that yield text the other. All of these operations
are described for synchronous context-free grammars in David Chiang's tutorial
\cite{Chiang06anintroduction}. In his tutorial, Chiang describes some of the
limitations of using synchronous CFGs; notably, they cannot rearrange parts of
parse trees that are not sisters. Of particular interest in this work is
raising and lowering elements; Chiang gives the example of swapping subjects
and objects, as in the example of translation between English and French in
Figure \ref{missesmary}. Chiang points out that, for syntax-aware MT, we would
like to be able to use some more powerful formalism that can perform
transformations like this. Synchronous tree substitution grammars, for example,
are able to describe transformations of this form, but not the transformation
from cross-serial dependencies in subordinate clauses in Dutch to the nested
clause structure of English. This latter transformation would require more
formal power, which is offered by tree-adjoining grammars.

\subsection{TAG and Related Formalisms}
\label{sec:tagfamily}
Tree adjoining grammar, introduced by Joshi
\cite{Joshi:1975:TAG:1739967.1740303},  has been a popular formalism for
describing grammars over trees. It provides additional expressive power not
available in regular tree grammars, handling some, but not all
context-sensitive languages. TAG can cleanly describe many of the
non-context-free features observed in human languages, such as the cross-serial
dependencies in Dutch. TAG is thus called ``weakly context-sensitive",
and has been shown formally equivalent to several other syntactic formalisms,
such as Combinatory Categorial Grammar (CCG) and Linear Indexed Grammars
\cite{vw94}.  

The operations of TAG are substitution and adjunction, which combine the two
different kinds of elementary trees present in a given TAG grammar, initial
trees and auxiliary trees. The substitution operation takes two trees, one with
a leaf that is an unresolved nonterminal $\alpha$, and produces a new tree in
which that node has been replaced with an entire subtree (copied from another
initial tree in the grammar, or one that has already been derived) whose root
node is also $\alpha$. For example, an initial tree may have an unresolved
nonterminal that wants to have an NP attached to it (it has a leaf labelled
NP); the substitution operation attaches an existing subtree whose root is NP,
producing a new tree where that nonterminal is now resolved.
The adjunction operation takes an existing tree and an auxiliary tree, which
has a special node marked as the ``foot", and grafts the auxiliary tree in
place in the middle of the existing tree, attaching the tree material at the
target location to the foot node of the auxiliary tree. For a very clear
tutorial on TAG with good examples, please see \cite{vannoord93}, Section
4.2.4. Synchronous TAG has also been investigated, and its use in machine
translation has been advocated by Shieber, who argues that its expressive power
may make up for its computational complexity \cite{textscshieber:2007:SSST}.

Restricted versions of TAG and their synchronous analogues have also been
investigated. These do not provide the full expressive power of TAG, but can be
parsed and trained more efficiently. The two limited versions of TAG that are
most prominently discussed in the literature are tree substitution grammars
(TSG) and tree insertion grammars (TIG). TSG only provides the substitution
operation, and does not have auxiliary trees or adjunction
\cite{eisner:2003:ACL-companion}. TIG, on the other hand, includes both the
substitution and adjunction operations, but places constraints on the
permissible shapes of auxiliary trees: their foot nodes must be at the leftmost
or rightmost edge of the frontier, and a given derivation may not adjoin
``left" auxiliary trees into ``right" ones, or vice-versa.  These restrictions
are sufficient to limit the weak generative capacity of TIGs to that of CFGs,
but they also ensure that algorithms on TIGs can run more efficiently.  While
parsing with a TAG takes in the general case $O(n^6)$ complexity, TIG (like the
general case for CFGs) can be parsed in $O(n^3)$ \cite{Nesson:2006:IPS}. Both
STIG and STSG have seen use in machine translation; for example, probabilistic
STIG is used in \cite{Nesson:2006:IPS}, and STSG has been notably used in
\cite{eisner:2003:ACL-companion}.

\subsection{Tree Transducers}
While synchronous grammars provide a \emph{declarative} description of a
relation that holds between two sets of trees, tree transducers more explicitly
describe the process by which a tree may be converted into other trees. Like
finite-state transducers, which operate over strings, tree transducers
typically describe nondeterministic processes, so for a given input tree, there
is a set of possible output trees; that set may (for example) be described by a
regular tree grammar.

Tree transducers and synchronous grammars both describe mathematical relations
over trees, so we can sensibly ask about their comparative formal expressive
power, and use them to compute similar queries. For example, with either a
synchronous grammar or a transducer, we may ask, for a given tree, what are the
other trees that are in the mathematical relation with it
\cite{Chiang06anintroduction}. There are transducer varieties with the same
formal expressive power as certain synchronous grammars.  For example,
synchronous tree substitution grammars (STSG) have the same formal power as
xLNT transducers \cite{Maletti:2010:WST:1857999.1858129}, which will be
described in more detail in the next section.

While there are very many kinds of possible tree transducers, the ones used in
NLP applications typically fall into one of two classes, T transducers, which
operate ``top-down", and ``B" transducers, which operate ``bottom-up".

\begin{figure*}
\begin{center}
\begin{itemize}
\item synchronous grammar: a grammar over two languages simultaneously, where
rules are given in pairs and must be used together
\item probabilistic grammar: a grammar where rules have associated weights,
which defines a probability distribution over derivations licensed by that
grammar
\item TAG: tree adjoining grammar, mildly context-sensitive grammar formalism
over trees, with substitution and adjunction operations
\item TIG: tree insertion grammar: a TAG wherein rules have certain
restrictions, described in Section \ref{sec:tagfamily}.
\item TSG: tree substitution grammar; similar to TAG without the adjoining
operation
\item RTG: regular tree grammar, the tree analogue of context-free grammar
\item finite-state transducers: transducers over strings; finite-state automata
with the added ability to produce output
\item tree transducers: automata that define relations over trees procedurally
\item T transducers: ``top down" tree transducers
\item R transducers: the same as T transducers, a name used in earlier work.
``R" stands for ``Root to Frontier"
\item (L)T transducers: ``linear" T transducers, constrained such that their
rules are non-copying, and a variable appearing on the left-hand side of a rule
must appear at most once in the right-hand side
\item (N)T transducers: ``nondeleting" T transducers, constrained such that
a variable appearing on the left-hand side of a rule must appear at least once
in the right-hand side
\item (x)T transducers: T transducers with ``extended" pattern matching,
allowing for complex finite patterns to appear in the left-hand side of
rules.
\end{itemize}
\end{center}
\caption{Glossary of transducers and automata}
\label{glossary}
\end{figure*}

\section{T Transducers}
Let us now describe T transducers in more detail. T transducers transform trees
into other trees via sets of production rules.
Many production rules may apply at a given step in a derivation, so the
transductions are usually nondeterministic, relating a given input tree to many
possible output trees.  Thus a T transducer, like a synchronous tree grammar,
defines a \emph{relation} over sets of trees.

\begin{figure*}
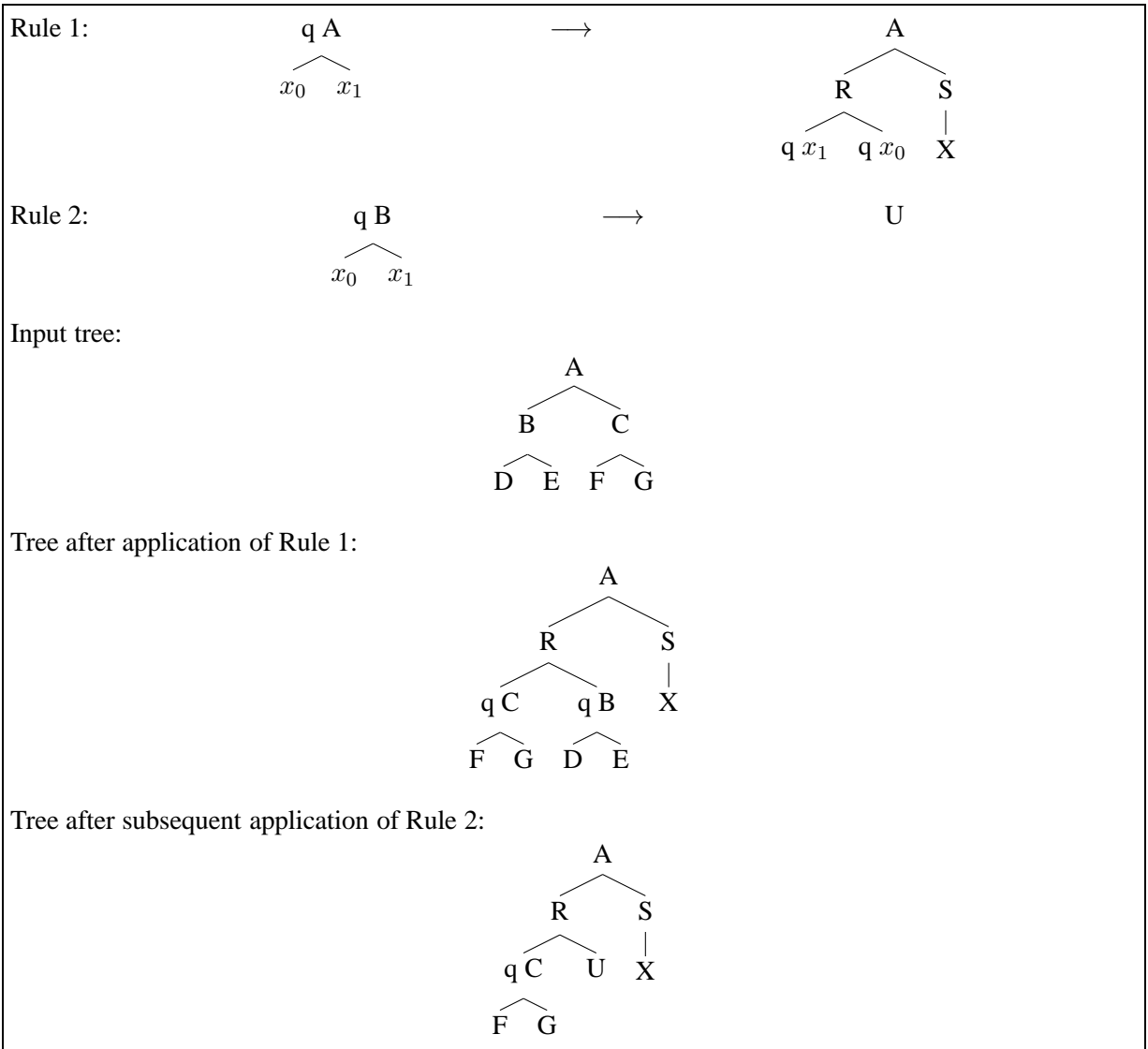

Rule 1:
\Tree [.{q A} $x_0$ $x_1$ ]
\hspace{1in}
$\longrightarrow$
\Tree [.A [.R {q $x_1$} {q $x_0$} ] [.S X ] ]

\bigskip
Rule 2:
\Tree [.{q B} $x_0$ $x_1$ ]
\hspace{1in}
$\longrightarrow$
\Tree [.U ]

\bigskip
Input tree:

\Tree [.A [.B D E ] [.C F G ] ]

\bigskip
Tree after application of Rule 1:

\Tree [.A [.R [.{q C} F G ] [.{q B} D E ] ] [.S X ] ]

\bigskip
Tree after subsequent application of Rule 2:

\Tree [.A [.R [.{q C} F G ] U ] [.S X ] ]

\caption{Example transduction steps, simplified from
\cite{DBLP:journals/coling/GraehlKM08}. Note that this transduction is not
complete because the node with the symbol ``C" is still in the state q.}
\label{exampletransduction}
\end{figure*}

Intuitively, transduction begins with an input tree, where its root node is in
the initial state $q_0$. Each node in a tree may be in one of the states in $Q$
(the set of possible states), or in no state at all. Transduction proceeds by
finding all the transduction rules that can apply to an existing tree, or
subtrees of an existing tree. A rule applies when the root of its left-hand
side matches a node in the tree, and the state of the node matches the state of
the rule.  When a rule matches a subtree (call it $t$), then a new tree is
produced and added to the set of current trees by replacing the subtree that
matched the rule with the right-hand side of the rule, save that the variables
in the right-hand side of the rule have been replaced by the corresponding
subtrees of $t$. Additionally, a rule may specify that subtrees of the new tree
being produced should be in states as well, indicating that more transduction
work must be done on them before the derivation is finished. A complete,
successful transduction in a T transducer begins with the root node being in
the initial state, then states propagating down the tree to the leaves, until
the entire tree has been transduced. See Figure \ref{exampletransduction} for
an illustrative example, adapted from \cite{DBLP:journals/coling/GraehlKM08}.

A new tree is produced by replacing the subtree that matched the rule with the
right-hand side of the rule, with its variables filled in with the appropriate
subtrees. The new tree is then added to the inventory of current trees in the
usual way for production systems. A transduction is complete for a tree in the
inventory when all of its nodes are no longer in states; at this point, the
states will have propagated all the way from the top of the tree to the leaves,
and then be resolved; in the case of translation, the symbols in the tree will
be words in the output language. The transduction process is nondeterministic;
many rules may apply to a given tree in the inventory, and even the same rule
may apply to different subtrees. To do a complete search for all possible
transductions, we apply each rule to every subtree where it is applicable, and
produce every possible resulting tree; beam search may also be done, where
search paths with low probabilities are pruned.

Formally, a T transducer has the following elements.

\begin{itemize}
\item an input alphabet $\Sigma$
\item an output alphabet $\Delta$
\item a set of \emph{states} $Q$
\item an \emph{initial state}, typically denoted $q_0$
\item transition rules, which are tuples of the form
$(q \subset Q,\sigma \subset \Sigma,tpat, p)$
\end{itemize}

The transition rule tuples specify the state that a given node must be in, and
the symbol from the input language that the subtree must have, (state $q$ and
symbol $\sigma$, respectively), in order for this rule to match. They also
specify a tree \emph{pattern} that forms the right-hand side of the rule, and a
weight $p$ for this rule. The tree pattern is a tree where some of the elements
in the tree may be variables, which refer to subtrees of the left-hand side
under consideration.

\subsection{xT Transducers}
\begin{figure*}
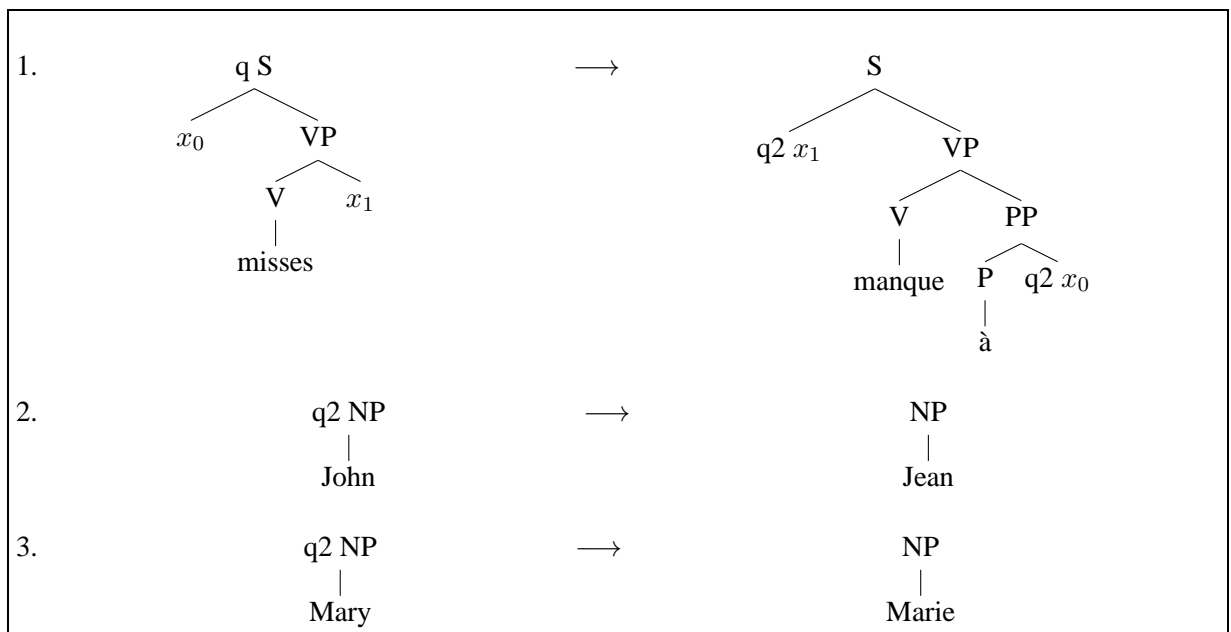

\bigskip
1. \Tree [.{q S} $x_0$ [.VP [.V misses ] $x_1$ ] ]
\hspace{1in}
$\longrightarrow$
\Tree [.S {q2 $x_1$} [.VP [.V manque ] [.PP [.P à ] {q2 $x_0$} ] ] ]

\bigskip
2. \Tree [.{q2 NP} John ]
\hspace{1in}
$\longrightarrow$
\Tree [.NP Jean ]

\bigskip
3. \Tree [.{q2 NP} Mary ]
\hspace{1in}
$\longrightarrow$
\Tree [.NP Marie ]
\caption{Switching subject and object in translation to French with an xT
transducer rule. The state q2 here indicates that we want to translate names as
well.} 
\label{missesanybody}
\end{figure*}

The ``extended" variation of T transducers, indicated with an ``x" prefix, adds
the capability for rules to check whether a potentially matching subtree
matches a certain pattern of finite size, in addition to the given state and
value of the node. The tree pattern in the left-hand side of an xT transduction
rule may contain literal symbols as well as variables, which allows for
lexicalized rules that only apply when certain words are in a subtree. The tree
patterns also make it possible for the rules to reference material finitely far
into a subtree, which makes local rotations straightforward; see Figure
\ref{missesanybody} for example xT rules that perform a local rotation and also
use finite lookahead to produce Francophone names. In the notation common in
the literature, a state for a node is written next to that node in the tree
structure.

While T transducers are not as expressive as synchronous TSG
\cite{Shieber:2004:SGT}, xT transducers are as expressive, and can
even be used to simulate synchronous TAG in some cases \cite{maletti:2010:ACL}.
In addition to their formal expressive power, xT transducers are much more
convenient for rule authors; some finite lookahead can be simulated with the
standard T transducers, as shown in \cite{DBLP:journals/coling/GraehlKM08}, but
it is somewhat tedious. The use of xT transducers makes writing rules to
rearrange material in a tree much more convenient.

\subsection{Restricted Versions of T and xT Transducers}
For computational efficiency purposes, we may also consider placing certain
restrictions on the rules in a T or xT transducer. Options that have been
explored include requiring that a transducer be \emph{linear}, which means that
any variable occurring in the left-hand side of a rule should appear no more
than once in the right-hand side, and \emph{nondeleting}, which means that a
variable in the left-hand side must appear at least once in the right-hand
side. Linear transducers are given the prefix ``L", and nondeleting transducers
the prefix ``N", so for example, extended linear non-deleting top-down
transducers are described as ``xLNT". This particular combination of options
has been used several times in the literature, including
\cite{galley-EtAl:2004:HLTNAACL}.  Also note that the transducer in Figure
\ref{exampletransduction} is not nondeleting, since Rule 2 does not reference
its variables in its right-hand side.

Among the benefits of adding these constraints on rules are that, LT and LNT
transducers are \emph{compositional}, meaning that a relation that can be
expressed by a cascade of two LT transducers can also be expressed by a single
LT transducer, and that the composition of those two transducers can be
computed. However this is not possible with any other members of the
T-transducer family; even xLNT transducers are non-compositional
\cite{DBLP:journals/mt/Knight07}.

\section{Linguistic Divergences}
Bonnie Dorr, in \cite{DBLP:journals/coling/Dorr94}, enumerates several
different kinds of structural divergences that we might see in translation
between languages.  These divergences occur when translating from English to
closely related languages, Spanish and German, all of which have fairly similar
word orders.  These are not the only kinds of syntactic differences
that there can be in a translation. They do not, for example, cover the more
large-scale reorderings that we see when translating between SVO and SOV or VSO
languages. However, each of these divergences require something more than
simple word substitution or reordering the children of a given node: many of
these require raising and lowering tree material (performing ``rotations",
in the terminology of \cite{Shieber:2004:SGT}), and nested phrases that are
present in one language are often not present in the other. Many of these
divergences may appear in a given pair of translated sentences. The following
subsections describe Dorr's seven kinds of divergence.

\subsection{Thematic Divergence}
Different languages may express a situation by assigning different thematic
roles to the participants of an action, swapping (for example) the subject and
object. For example, translating from English to Spanish, we see:

\begin{itemize}
\item I like Mary
\item María me gusta a mí
\end{itemize}

In Spanish it is more common to say that ``X pleases Y" than that "Y
wants/likes X". The Spanish verb \textit{querer} has the same structure as the
English ``like", but the meaning of ``gustar" is closer to the English ``to
like".

\subsection{Promotional Divergence}
A modifier in one language may be the head in another language.

\begin{itemize}
\item John usually goes home
\item Juan suele ir a casa
\end{itemize}

Here in English, an adverb modifies the verb to indicate that it is habitual,
but in Spanish we use the verb ``soler" (which inflects as ``suele" for
third-person singular), to express this. It has an infinitive as a dependent.

\subsection{Demotional Divergence}
The demotional divergence is similar to a demotional divergence, viewed in the
other direction; in cases of demotional divergence, a head in one language is a
modifier in the other. In \cite{DBLP:journals/coling/Dorr94}, a formal
distinction is made between the two because in Dorr's MT system, they would be
triggered in different circumstances, but for our purposes they are
effectively analogous.

\begin{itemize}
\item I like eating
\item Ich esse gern
\end{itemize}

In this example, while English uses the verb ``to like", German has an adverb.
The sentence has a literal translation of ``I eat likingly".

\subsection{Categorial Divergence}

In cases of categorial divergence, the meaning of a word with a certain part of
speech in one language is expressed with a different part of speech in the
other.

\begin{itemize}
\item I am hungry
\item Ich habe Hunger.
\end{itemize}

The German sentence here translates literally as ``I have hunger."

\subsection{Structural Divergence}

In cases of structural divergence, there are phrases in one language not
present in the other.

\begin{itemize}
\item John entered the house
\item Juan entró en la casa
\end{itemize}

While the English sentence has the destination of the motion verb as an object,
in the Spanish we see the prepositional phrase ``en la casa" (``in the house").

\subsection{Lexical Divergence}
In cases of lexical divergence, the two languages involved have different
idiomatic phrases for describing a situation.

\begin{itemize}
\item John broke into the room 
\item Juan forzó la entrada al cuarto
\end{itemize}

While ``break into" is a phrasal verb in English, in Spanish it is more
idiomatic to ``force entry to". This example also includes a structural
divergence, as ``al cuarto" is a prepositional phrase not present in the
English.

\subsection{Conflational Divergence}
The meaning of the sentence may be distributed to different words in a
different language; the meaning of a verb, for example, may be carried by a
verb and its object after translation.

\begin{itemize}
\item I stabbed John
\item Yo le di puñaladas a Juan
\end{itemize}

Here the Spanish sentence means literally ``I gave John knife wounds". The words
``le" and ``a" are both required, but for different reasons: the verb ``dar"
(to give) requires the personal pronoun beforehand, and whenever a human being
is the object of a verb in Spanish, we add the ``personal a" beforehand.

\section{Implementation}
In the course of this project, I have produced a small, easily-understandable
toolkit named \texttt{kurt} (the Keen Utility for Rewriting Trees), for
experimenting with weighted xT tree transducers. It is implemented in Python 3
and makes use of the NLTK tree libraries \cite{nltkbook}. \texttt{kurt} has
been released as free software, and is available online
\footnote{http://github.com/alexrudnick/kurt}.

The software can perform tree transduction in general for weighted xT
transducers: given a tree, it applies xT transduction rules and produces a list
of output trees.  The implementation is fairly naïve, and proceeds as a simple
production system. Partial solutions are matched against every rule in the
transducer, then each matching rule is applied to the partial solution,
producing a new generation of partial solutions. Eventually, the derivation
either succeeds by producing at least one tree with no nodes in a state, or
it fails if the input tree cannot be completely transduced by the given rules.
The system returns all possible output trees, and the complete solutions are
printed out at the conclusion of the program.

The xT rules are straightforward to write, and are stored in YAML files. I have
also provided example xT rules that translate the examples of divergences given
by Dorr; these are described in more detail in Section
\ref{sec:dorr-transducers}.

A complete and useful MT system based on this software -- such that the rules
and their weights were not completely the product of human knowledge
engineering -- would require the implementation of a few more algorithms
described in \cite{DBLP:journals/coling/GraehlKM08}, particularly their EM
training algorithm to calculate weights for a given set of transduction rules,
which depends on their transduction algorithm that produces the more compact
representation of a transduction, a RTG. Decoding would require beam search
over tree transduction, or perhaps over generation using this compact RTG
representation. Additionally, some clever algorithm for extracting tree
transducer rules from parallel treebanks would be useful for the case where
parallel treebanks are available; some candidate techniques for this last
problem are discussed in Section \ref{sec:extraction}.

\subsection{Using the Software}
Transducers are stored in YAML files, with one xT transducer per file; each
rule is specified as an entry in that YAML file, and contains the following
entries.

\begin{itemize}
\item \texttt{state}: (required) The name of the state that a node at the root
of a subtree must be in to match this rule
\item \texttt{lhs}: (required) The left-hand side of the rule: a tree pattern,
typically with variables (tokens starting with \texttt{?}) that must unify with a
subtree in order for that subtree to match this rule
\item \texttt{rhs}: (required) The right-hand side of the rule: another tree
pattern, which is filled in when this rule is applied. It may contain
variables, in which case all of the variables must also be present in the
left-hand side of the rule.
\item \texttt{newstates}: (optional) Specifies the locations of transduction
states in the subtree produced by this rule. There may be many states specified
in the new subtree. They are given in the form
\texttt{[location, statename]}, where location is a bracketed list that
describes the path down the tree from the root of the subtree, with 0-indexed
children. For example, to put the second child of the leftmost child of the
root in state \texttt{foo}, a rule would have a \texttt{newstates} member
\texttt{[[0,1], foo]}.
\item \texttt{weight}: (optional) The weight for this rule. If unspecified, it
defaults to 1.0.
\end{itemize}

Given a file with these entries for each rule of a transducer, say called
\texttt{translation.yaml}, a Python 3 program can use \texttt{kurt} to do tree
transductions in the following way, assuming the libraries are all in the
\texttt{\$PYTHONPATH} or the current working directory.

\begin{verbatim}
from loadrules import loadrules
from translate import translate

rules = loadrules("translation.yaml")
tr = Tree("""(S (NP (PRP I))
                (VP (VB am)
                    (JJ hungry)))""")

## print all valid transductions
translate(tr, rules)
\end{verbatim}

\subsection{Simple Topicalization Example}
In Figure \ref{topicalization}, we see a toy example of xT rules realized with
the system. This is a complete running example that exercises many features of
the software; it translates an English sentence into ``LOLcat" Internet
slang, which features more prominent topicalization \footnote{Readers may or
may not be familiar with the moderately popular catchphrase ``My Pokemans, let
me show you them".}. For simplicity, the syntactic structure of the parse tree
is elided. The initial rule matches a sentence in the initial state \texttt{q},
containing ``let me show you my $x_0$" and produces a new sentence where ``my
$x_0$" has been moved to the front . The rule also specifies that the
(0-indexed) child of the S node at index $1$ is in the state \texttt{respell}.
The second rule matches the word ``Pokémon" when it is in the state
\texttt{respell}, replacing it with the slang spelling of ``Pokemans". The
third rule is for generalization, allowing words other than ``Pokémon" to be
translated in this position. Due to both the second and third rules applying to
the subtree, both spellings are produced in the output, but the translation
with the slang spelling is given a higher weight.

\begin{figure*}
\begin{center}
\begin{verbatim}
## lolcat topicalization (fronting)
- state: q
  lhs: (S let me show you my ?x0)
  rhs: (S my ?x0 , let me show you them)
  newstates:
  - [[1], respell]
- state: respell
  lhs: Pokémon
  rhs: Pokemans
  weight: 0.9
- state: respell
  lhs: ?x0
  rhs: ?x0
  weight: 0.1
\end{verbatim}
\end{center}
\caption{xT rules for translating into LOLcat dialect, which features
topicalization, in the YAML format used by the software implemented as part of
this work}
\label{topicalization}
\end{figure*}

\section{xT Transducers for Linguistic Divergences}
\label{sec:dorr-transducers}

I wrote xT transduction rules for the software toolkit that handle each of
Dorr's divergence examples. Most of the work involved was constructing parse
trees for the source- and target-language sentences; I then converted the trees
into templates for the desired trees, at which point they were effectively xT
transduction rules.  Some examples are included in Figures
\ref{translationGerman} and \ref{translationSpanish}, but the complete set of
rules are in \texttt{german.yaml} and \texttt{spanish.yaml}, included with the
software.  Most of the transformations required to implement these rules are
instances of local rotations, as described by \cite{Shieber:2004:SGT}. 

\begin{figure*}
\begin{center}
\begin{verbatim}
# handle <pronoun> like <gerund>
- state: q
  lhs: (S (NP ?x0) (VP (VB like) ?x1))
  rhs: (S (NP ?x0) (VP ?x1 (RB gern)))
  newstates: 
  - [[0, 0], lookup]
  - [[1, 0], gerundtotensed]

# handle I am <adj>
- state: q
  lhs: (S (NP ?x0) (VP (VB am) ?x1))
  rhs: (S (NP ?x0) (VP (VB habe) ?x1))
  newstates: 
  - [[0, 0], lookup]
  - [[1, 1], adjtonoun]

## simple lookups for known phrases
- state: lookup
  lhs: (PRP I)
  rhs: (PRP ich)

## POS changes.
- state: gerundtotensed
  lhs: (VBG eating)
  rhs: (VB esse)

- state: adjtonoun
  lhs: (JJ hungry)
  rhs: (NN hunger)
\end{verbatim}
\end{center}
\caption{Sample translation rules for German}
\label{translationGerman}
\end{figure*}

\begin{figure*}
\begin{center}
\begin{verbatim}
# handle <pronoun> like <name>
- state: q
  lhs: (S (NP ?x0) (VP (VB like) (NP ?x1)))
  rhs: (S (NP ?x1) (VP (NP ?x0) (VB gusta) ?x0))
  newstates: 
  - [[0, 0], lookup]
  - [[1, 0, 0], objectivize]
  - [[1, 2], tothisperson]
- state: tothisperson
  lhs: (PRP I)
  rhs: (PP (A a) (PRP mí))
# handle usually -> soler
- state: q
  lhs: (S (NP ?x0) (VP (RB usually) ?x1))
  rhs: (S (NP ?x0) (VP (VBZ suele) ?x1))
  newstates: 
  - [[0,0], lookup]
  - [[1,1], unconjugate]
# handle entered-object -> entró en ...
- state: q
  lhs: (S (NP ?x0) (VP (VBD entered) ?x1))
  rhs: (S (NP ?x0) (VP (VBD entró) (PP (IN en) ?x1)))
  newstates: 
  - [[0,0], lookup]
  - [[1,1,1], lookup]
# handle broke-into X -> forzó la entrada a X
- state: q
  lhs: (S (NP ?x0) (VP (VBD broke) (PP (IN into) ?x1)))
  rhs: (S (NP ?x0) (VP (VBD forzó)
                       (NP (DT la) (NN entrada) (PP (IN a) ?x1))))
  newstates: 
  - [[0,0], lookup]
  - [[1,1,2,1], lookup]
  - [[1,1,2], al]
# handle I stabbed X -> le di puńaladas a 
- state: q
  lhs: (S (NP ?x0) (VP (VBD stabbed) ?x1))
  rhs: (S (NP ?x0) (VP (PRP le) (VBD di)
                       (NP (NN puñaladas)) (NP (A a) (NNP Juan))))
  newstates: 
  - [[0,0], lookup]
\end{verbatim}
\end{center}
\caption{Sample translation rules for Spanish}
\label{translationSpanish}
\end{figure*}

\section{Related Work}
\label{sec:relatedwork}

In addition to the work on tree-based MT, some very sophisticated
string-based MT algorithms have been framed in terms of finite-state
transducers. Not long after the introduction of modern word-based SMT, Knight
and Al-Onaizan showed that IBM Model 3 could be expressed with a cascade of
FSTs \cite{DBLP:conf/amta/KnightA98}. Since string transducers can be composed,
decoding in this case becomes one enormous beam search over a single state
machine. Similarly, Shankar Kumar and William Byrne expressed the phrase-based 
alignment template model as FSTs \cite{DBLP:conf/naacl/KumarB03}. The last part
of the decoding process in Chiang's hierarchical phrase-based model can also be
described in terms of FSTs \cite{iglesias-EtAl:2009:NAACLHLT09}; Iglesias et
al. use finite-state techniques to traverse a lattice of possible translations
once chart parsing with an SCFG has completed.

For tutorials and related algorithms, Chiang provides an excellent introduction
to synchronous grammars in \cite{Chiang06anintroduction}. My understanding of
TAG was greatly aided by the TAG section in \cite{vannoord93}; it is referenced
in the TAG Wikipedia page. For overviews of different applications of T-family
tree transducers and their various properties, in a very approachable style,
\cite{DBLP:journals/mt/Knight07} and \cite{KnightGraehlOverview} are very
helpful. Additionally \cite{DBLP:journals/coling/GraehlKM08} contains excellent
examples for understanding xT transduction (one of which is in this paper in
simplified form, though the original example is worth working through and
understanding fully), along with a set of algorithms that can be computed over
xT transducers, including an EM procedure that can be used to estimate the
weights for an xT grammar given a parallel treebank.

\section{Conclusions and Future Work}
Here I have described the ``T" family of tree transducers and situated them
among the various formalisms for describing relations over strings and trees;
I have also demonstrated that xT transducers are sufficient for handling
translation across the linguistic divergences described by Dorr.  I have
presented a software package suitable for experimentation with xT transducers,
which comes with example translation rules that perform translations over each
of the divergences.

There remains significant work to be done on the topic; for example, to my
knowledge, there is no easily available end-to-end MT system based on tree
transducers, either commercial or Open Source. There are many more questions
that I would like to answer; as far as I know, these are open problems in the
field.

\subsection{Transducers, Disambiguation, and Language Models}
While weighted synchronous grammars and xT transducers provide generative
models of translation, the probabilities that they assign to a given rule are
set ahead of time, and are not conditioned on features of the surrounding
context. It may be fruitful to try using discriminative approaches (i.e.,
classifiers) to help a transducer-based MT system make decisions about which
rules are the most likely to apply in a given context, either based on the
surrounding tree material, or on the surface words in the source-language
sentence. It may turn out that there is a more principled way to achieve the
same benefits, perhaps by adding more conditions on the probabilities in a
generative model. However, cross-language phrase-sense disambiguation with
classifiers, like in the work of Carpuat and Wu \cite{carpuatpsd}, has proved
useful for phrase-based SMT. For phrase-based SMT in general, discriminative
approaches such as Minimum Error-Rate Training (MERT)
\cite{DBLP:conf/acl/Och03} have become quite typical.

Another guide for the tree transduction process could be language models,
either flat n-gram models or structured ones, which would have the added
benefit that they could be trained on larger corpora than those used to produce
the tree transduction rules in the first place.

\subsection{Extraction and Training Transducers}
\label{sec:extraction}
Thus far, it seems as though there is no agreed-upon best approach for
extracting a set of tree transduction rules from a parallel
treebank, such that a tree-to-tree MT system could be constructed.  While
parallel treebanks are not abundant, with sufficiently good monolingual
parsers, parallel trees can be created from bitext, and hopefully these could
be used to induce transduction rules for tree-to-tree MT systems. Other work
has presented methods for learning tree-to-string transduction rules, for
example \cite{galley-EtAl:2004:HLTNAACL} and \cite{deneefe-knight:2009:EMNLP}.
These approaches for learning tree-to-string transducers, if I understood them
more completely, might turn out to generalize easily to the tree-to-tree case,
but if so, it is not yet obvious to me how to do this.

One proposed approach for learning relations over trees is given in
\cite{eisner:2003:ACL-companion}, in which Eisner presents algorithms for both
extracting an STSG grammar and training its weights; STSGs can then be
expressed as xT transducers as described by Maletti in \cite{maletti:2010:ACL}.
Additionally, approaches for leaning tree transduction rules have been
suggested for tasks other than machine translation, particularly in the
summarization work of Cohn and Lapata \cite{cohn-lapata:2007:EMNLP-CoNLL2007},
\cite{cohn-lapata:2008:PAPERS}, who work with a corpus that not only has parse
trees for both source and target languages (in their case, pairs of longer and
paraphrased sentences, both in English), but has also been word-aligned. The
word alignments inform their grammar extraction. Cohn and Lapata use a very
small training paraphrase corpus (480 sentences), which suggests that perhaps
their methods would be useful for MT with low-resourced languages. They also
use of discriminative methods for training and decoding. Both their algorithm
for rule extraction and the tree transducers with discriminative methods may
have been used in tree-to-tree MT system, but I have not yet found work that
describes this; if it has not yet been tried, someone should explore it.

\subsection{XDG as Transducers}
Given that many grammar formalisms are expressible in terms of tree
transducers, one wonders if constraint-based dependency frameworks, such as
Extensible Dependency Grammar \cite{Debusmann06}, which has been used by
Michael Gasser for machine translation \cite{gasser:2011:freerbmt}, could be
expressed in terms of tree transducers. Transducers over dependency trees have
already been used for machine translation, for example by Ding and Palmer
\cite{ding-palmer:2005:ACL}. However, XDG defines not just one layer of
dependency analysis for a language, but several. Its analysis of a sentence
in a given language is a multigraph with multiple dimensions of analysis, with
constraints describing permissible structures on each dimension, as well as the
relationships between dimensions. This suggests that perhaps XDG could be
expressed as a cascade of transducers, with each layer in the cascade
describing the relation between one XDG dimension and the next.

A problem with this interpretation is that not all layers of an XDG multigraph
are tree structures. This might mean that XDG cannot be cleanly expressed in
this way at all, or perhaps that another kind of transducer that operates on
graphs more generally could be used. Alternatively, perhaps XDG could be tweaked
such that every layer has a tree structure.

If it is in fact possible to express XDG translation rules as a cascade of
transducers, then this would present a clear path for integrating machine
learning into the largely rule-based system, making use of the training
algorithms already present in the literature. As a fairly modest step, given
small numbers of parallel training sentences, one could use EM to train the
weights of the transduction rules that implement the XDG grammar. More
ambitiously, one could perhaps extract grammar rules from example translation
pairs, although the XDG parse graphs would have to be provided by an expert,
for each layer in the analysis. This could be done either simply on demand,
when the existing grammar fails to parse and translate a sentence, or using
active learning to select sentences for human annotation.

One problem not addressed at all in the literature that I have seen is how to
translate, either into or out of, morphologically rich languages using tree
transducers. It seems as though morphological analysis and lemmatization would
be an important first step in a transducer-based MT system, to limit the number
of rules that the system needs to consider, but then the morphological
information should be used to help the system make choices during transduction
(decoding). Perhaps morphological features would be useful to classifiers
trained to help make syntactic disambiguation decisions.

\bibliographystyle{acl}
\bibliography{transducers.bib}

\begin{thebibliography}{}

\bibitem[\protect\citename{Allauzen \bgroup et al.\egroup }2007]{openfst}
Cyril Allauzen, Michael Riley, Johan Schalkwyk, Wojciech Skut, and Mehryar
  Mohri.
\newblock 2007.
\newblock Open{F}st: A general and efficient weighted finite-state transducer
  library.
\newblock In {\em Proceedings of the Ninth International Conference on
  Implementation and Application of Automata, (CIAA 2007)}, volume 4783 of {\em
  Lecture Notes in Computer Science}, pages 11--23. Springer.
\newblock {\tt http://www.openfst.org}.

\bibitem[\protect\citename{Bird \bgroup et al.\egroup }2009]{nltkbook}
Steven Bird, Ewan Klein, and Edward Loper.
\newblock 2009.
\newblock {\em {Natural Language Processing with Python}}.
\newblock O'Reilly Media.

\bibitem[\protect\citename{Brown \bgroup et al.\egroup
  }1993]{DBLP:journals/coling/BrownPPM94}
Peter~F. Brown, Stephen~Della Pietra, Vincent J.~Della Pietra, and Robert~L.
  Mercer.
\newblock 1993.
\newblock The mathematics of statistical machine translation: Parameter
  estimation.
\newblock {\em Computational Linguistics}, 19(2):263--311.

\bibitem[\protect\citename{Carpuat and Wu}2007]{carpuatpsd}
Marine Carpuat and Dekai Wu.
\newblock 2007.
\newblock How phrase sense disambiguation outperforms word sense disambiguation
  for statistical machine translation.
\newblock In {\em 11th Conference on Theoretical and Methodological Issues in
  Machine Translation}.

\bibitem[\protect\citename{Chiang}2005]{chiang:2005:ACL}
David Chiang.
\newblock 2005.
\newblock A hierarchical phrase-based model for statistical machine
  translation.
\newblock In {\em Proceedings of the 43rd Annual Meeting of the Association for
  Computational Linguistics (ACL'05)}, pages 263--270, Ann Arbor, Michigan,
  June. Association for Computational Linguistics.

\bibitem[\protect\citename{Chiang}2006]{Chiang06anintroduction}
David Chiang.
\newblock 2006.
\newblock An introduction to synchronous grammars.

\bibitem[\protect\citename{Cohn and
  Lapata}2007]{cohn-lapata:2007:EMNLP-CoNLL2007}
Trevor Cohn and Mirella Lapata.
\newblock 2007.
\newblock Large margin synchronous generation and its application to sentence
  compression.
\newblock In {\em Proceedings of the 2007 Joint Conference on Empirical Methods
  in Natural Language Processing and Computational Natural Language Learning
  (EMNLP-CoNLL)}, pages 73--82, Prague, Czech Republic, June. Association for
  Computational Linguistics.

\bibitem[\protect\citename{Cohn and Lapata}2008]{cohn-lapata:2008:PAPERS}
Trevor Cohn and Mirella Lapata.
\newblock 2008.
\newblock Sentence compression beyond word deletion.
\newblock In {\em Proceedings of the 22nd International Conference on
  Computational Linguistics (Coling 2008)}, pages 137--144, Manchester, UK,
  August. Coling 2008 Organizing Committee.

\bibitem[\protect\citename{Debusmann}2006]{Debusmann06}
Ralph Debusmann.
\newblock 2006.
\newblock {\em Extensible Dependency Grammar: A Modular Grammar Formalism Based
  On Multigraph Description}.
\newblock {Ph.D.} thesis, Saarland University, 4.

\bibitem[\protect\citename{DeNeefe and Knight}2009]{deneefe-knight:2009:EMNLP}
Steve DeNeefe and Kevin Knight.
\newblock 2009.
\newblock Synchronous tree adjoining machine translation.
\newblock In {\em Proceedings of the 2009 Conference on Empirical Methods in
  Natural Language Processing}, pages 727--736, Singapore, August. Association
  for Computational Linguistics.

\bibitem[\protect\citename{Ding and Palmer}2005]{ding-palmer:2005:ACL}
Yuan Ding and Martha Palmer.
\newblock 2005.
\newblock Machine translation using probabilistic synchronous dependency
  insertion grammars.
\newblock In {\em Proceedings of the 43rd Annual Meeting of the Association for
  Computational Linguistics (ACL'05)}, pages 541--548, Ann Arbor, Michigan,
  June. Association for Computational Linguistics.

\bibitem[\protect\citename{Dorr}1994]{DBLP:journals/coling/Dorr94}
Bonnie~J. Dorr.
\newblock 1994.
\newblock Machine translation divergences: A formal description and proposed
  solution.
\newblock {\em Computational Linguistics}, 20(4):597--633.

\bibitem[\protect\citename{Eisner}2003]{eisner:2003:ACL-companion}
Jason Eisner.
\newblock 2003.
\newblock Learning non-isomorphic tree mappings for machine translation.
\newblock In {\em The Companion Volume to the Proceedings of 41st Annual
  Meeting of the Association for Computational Linguistics}, pages 205--208,
  Sapporo, Japan, July. Association for Computational Linguistics.

\bibitem[\protect\citename{Galley \bgroup et al.\egroup
  }2004]{galley-EtAl:2004:HLTNAACL}
Michel Galley, Mark Hopkins, Kevin Knight, and Daniel Marcu.
\newblock 2004.
\newblock What's in a translation rule?
\newblock In Daniel~Marcu Susan~Dumais and Salim Roukos, editors, {\em
  HLT-NAACL 2004: Main Proceedings}, pages 273--280, Boston, Massachusetts,
  USA, May 2 - May 7. Association for Computational Linguistics.

\bibitem[\protect\citename{Gasser}2011]{gasser:2011:freerbmt}
Michael Gasser.
\newblock 2011.
\newblock Toward synchronous extensible dependency grammar.
\newblock In {\em Proceedings of the International Workshop on Free/Open-Source
  Rule-Based Machine Translation (2nd : 2011 : Barcelona)}, Barcelona,Spain,
  January.

\bibitem[\protect\citename{Graehl \bgroup et al.\egroup
  }2008]{DBLP:journals/coling/GraehlKM08}
Jonathan Graehl, Kevin Knight, and Jonathan May.
\newblock 2008.
\newblock Training tree transducers.
\newblock {\em Computational Linguistics}, 34(3):391--427.

\bibitem[\protect\citename{Iglesias \bgroup et al.\egroup
  }2009]{iglesias-EtAl:2009:NAACLHLT09}
Gonzalo Iglesias, Adri\`{a} de~Gispert, Eduardo R.~Banga, and William Byrne.
\newblock 2009.
\newblock Hierarchical phrase-based translation with weighted finite state
  transducers.
\newblock In {\em Proceedings of Human Language Technologies: The 2009 Annual
  Conference of the North American Chapter of the Association for Computational
  Linguistics}, pages 433--441, Boulder, Colorado, June. Association for
  Computational Linguistics.

\bibitem[\protect\citename{Joshi \bgroup et al.\egroup
  }1975]{Joshi:1975:TAG:1739967.1740303}
Aravind~K. Joshi, Leon~S. Levy, and Masako Takahashi.
\newblock 1975.
\newblock Tree adjunct grammars.
\newblock {\em J. Comput. Syst. Sci.}, 10(1):136--163, February.

\bibitem[\protect\citename{Kanthak and Ney}2004]{kanthak-ney:2004:ACL}
Stephan Kanthak and Hermann Ney.
\newblock 2004.
\newblock Fsa: An efficient and flexible c++ toolkit for finite state automata
  using on-demand computation.
\newblock In {\em Proceedings of the 42nd Meeting of the Association for
  Computational Linguistics (ACL'04), Main Volume}, pages 510--517, Barcelona,
  Spain, July.

\bibitem[\protect\citename{Knight and
  Al-Onaizan}1998]{DBLP:conf/amta/KnightA98}
Kevin Knight and Yaser Al-Onaizan.
\newblock 1998.
\newblock Translation with finite-state devices.
\newblock In David Farwell, Laurie Gerber, and Eduard~H. Hovy, editors, {\em
  AMTA}, volume 1529 of {\em Lecture Notes in Computer Science}, pages
  421--437. Springer.

\bibitem[\protect\citename{Knight and Graehl}2005]{KnightGraehlOverview}
Kevin Knight and Jonathan Graehl.
\newblock 2005.
\newblock An overview of probabilistic tree transducers for natural language
  processing.
\newblock In Alexander Gelbukh, editor, {\em Computational Linguistics and
  Intelligent Text Processing}, volume 3406 of {\em Lecture Notes in Computer
  Science}, pages 1--24. Springer Berlin / Heidelberg.

\bibitem[\protect\citename{Knight}2007]{DBLP:journals/mt/Knight07}
Kevin Knight.
\newblock 2007.
\newblock Capturing practical natural language transformations.
\newblock {\em Machine Translation}, 21(2):121--133.

\bibitem[\protect\citename{Kumar and Byrne}2003]{DBLP:conf/naacl/KumarB03}
Shankar Kumar and William~J. Byrne.
\newblock 2003.
\newblock A weighted finite state transducer implementation of the alignment
  template model for statistical machine translation.
\newblock In {\em HLT-NAACL}.

\bibitem[\protect\citename{Maletti}2010a]{maletti:2010:ACL}
Andreas Maletti.
\newblock 2010a.
\newblock A tree transducer model for synchronous tree-adjoining grammars.
\newblock In {\em Proceedings of the 48th Annual Meeting of the Association for
  Computational Linguistics}, pages 1067--1076, Uppsala, Sweden, July.
  Association for Computational Linguistics.

\bibitem[\protect\citename{Maletti}2010b]{Maletti:2010:WST:1857999.1858129}
Andreas Maletti.
\newblock 2010b.
\newblock Why synchronous tree substitution grammars?
\newblock In {\em Human Language Technologies: The 2010 Annual Conference of
  the North American Chapter of the Association for Computational Linguistics},
  HLT '10, pages 876--884, Stroudsburg, PA, USA. Association for Computational
  Linguistics.

\bibitem[\protect\citename{Nesson \bgroup et al.\egroup }2006]{Nesson:2006:IPS}
Rebecca Nesson, Stuart~M. Shieber, and Alexander Rush.
\newblock 2006.
\newblock Induction of probabilistic synchronous tree-insertion grammars for
  machine translation.
\newblock In {\em Proceedings of the 7th Conference of the {Association for
  Machine Translation in the Americas} ({AMTA} 2006)}, Boston, Massachusetts,
  8-12 August.

\bibitem[\protect\citename{Och and Ney}2004]{DBLP:journals/coling/OchN04}
Franz~Josef Och and Hermann Ney.
\newblock 2004.
\newblock The alignment template approach to statistical machine translation.
\newblock {\em Computational Linguistics}, 30(4):417--449.

\bibitem[\protect\citename{Och}2003]{DBLP:conf/acl/Och03}
Franz~Josef Och.
\newblock 2003.
\newblock Minimum error rate training in statistical machine translation.
\newblock In Erhard~W. Hinrichs and Dan Roth, editors, {\em ACL}, pages
  160--167. ACL.

\bibitem[\protect\citename{Shieber}2004]{Shieber:2004:SGT}
Stuart~M. Shieber.
\newblock 2004.
\newblock Synchronous grammars as tree transducers.
\newblock In {\em Proceedings of the Seventh International Workshop on Tree
  Adjoining Grammar and Related Formalisms (TAG+ 7)}, Vancouver, Canada, May
  20-22.

\bibitem[\protect\citename{Shieber}2007]{textscshieber:2007:SSST}
Stuart~M. Shieber.
\newblock 2007.
\newblock Probabilistic synchronous tree-adjoining grammars for machine
  translation: The argument from bilingual dictionaries.
\newblock In {\em Proceedings of SSST, NAACL-HLT 2007 / AMTA Workshop on Syntax
  and Structure in Statistical Translation}, pages 88--95, Rochester, New York,
  April. Association for Computational Linguistics.

\bibitem[\protect\citename{van Noord}1993]{vannoord93}
Gertjan van Noord.
\newblock 1993.
\newblock {\em Reversibility in Natural Language Processing}.
\newblock {Ph.D.} thesis, University of Utrecht.

\bibitem[\protect\citename{Vijay-Shanker and Weir}1994]{vw94}
K.~Vijay-Shanker and David Weir.
\newblock 1994.
\newblock The equivalence of four extensions of context-free grammars.
\newblock {\em Mathematical Systems Theory}, 27:511--546.

\end{thebibliography}
\end{document}